\newcommand{\CLASSINPUTtoptextmargin}{19mm}
\newcommand{\CLASSINPUTbottomtextmargin}{18mm}
\newcommand{\CLASSINPUTinnersidemargin}{19mm}
\newcommand{\CLASSINPUToutersidemargin}{19mm}

\documentclass[letterpaper,conference,final]{IEEEtran}
\IEEEoverridecommandlockouts
\usepackage{cite}
\usepackage{amsmath,amssymb,amsfonts,color}
\usepackage{algorithm}
\usepackage[noend]{algpseudocode}
\usepackage{placeins}
\usepackage{graphicx}
\usepackage{textcomp}
\usepackage{multicol,comment}
\usepackage{soul}

\def\BibTeX{{\rm B\kern-.05em{\sc i\kern-.025em b}\kern-.08em
    T\kern-.1667em\lower.7ex\hbox{E}\kern-.125emX}}
\begin{document}

\title{\vspace{6mm}\bf \LARGE Decentralized Multi-Floor Exploration by a Swarm of Miniature Robots Teaming with Wall-Climbing Units
\thanks{The authors acknowledge the support from the TL@SUTD-Systems Technology for Autonomous Reconnaissance \& Surveillance, the SUTD-MIT International Design Center (http://idc.sutd.edu.sg), and an MOE Tier 1 (Grant \#T1MOE1701)}
}

\author{Jabez L. Kit$^{1}$, Audelia G. Dharmawan$^1$, David Mateo$^1$, Shaohui Foong$^1$, Gim Song Soh$^1$,\\ Roland Bouffanais$^{1}$, and Kristin L. Wood$^1$
  \thanks{$^{1}$These authors are with the Singapore
    University of Technology and Design, Singapore 487372, e-mail: ({\tt\footnotesize bouffanais@sutd.edu.sg})}
}

\maketitle

\begin{abstract}
In this paper, we consider the problem of collectively exploring unknown and dynamic environments with a decentralized heterogeneous multi-robot system consisting of multiple units of two variants of a miniature robot. The first variant---a wheeled ground unit---is at the core of a swarm of floor-mapping robots exhibiting scalability, robustness and flexibility. These properties are systematically tested and quantitatively evaluated in unstructured and dynamic environments, in the absence of any supporting infrastructure. The results of repeated sets of experiments show a consistent performance for all three features, as well as the possibility to inject units into the system while it is operating. Several units of the second variant---a wheg-based wall-climbing unit---are used to support the swarm of mapping robots when simultaneously exploring multiple floors by expanding the distributed communication channel necessary for the coordinated behavior among platforms. Although the occupancy-grid maps obtained can be large, they are fully distributed. Not a single robotic unit possesses the overall map, which is not required by our cooperative path-planning strategy.
\end{abstract}


\section{Introduction}

Autonomous robots are good contestants for intelligent surveillance and reconnaissance (ISR) operations in remote or hazardous environments preventing direct human intervention. One central challenge in ISR operations, however, is the ability to perform effective exploration of dynamic environments. Single-robot autonomous exploration is unquestionably ill-suited for such tasks. Coordinated multi-robot exploration represents, theoretically at least, a viable alternative~\cite{burgard2005coordinated}.

Biological multi-agent systems---e.g. bird flocks, schools of fish, ant colonies---are capable of performing a wide range of collective behaviors in a fully decentralized manner~\cite{B2}. These swarming systems present valuable insights into the development of decentralized, scalable and fault-tolerant multi-robot systems (MRS) that are required to operate in dynamic environments. Swarm-based designs of MRS therefore appear to be a promising strategy for ISR operations in dynamic and unknown environments. However, as highlighted in~\cite{rone2013mapping}, the transition from a robot-centric design to a system-centric one requires to consider critical elements beyond the electromechanical aspects at the robotic unit level.

Decentralized MRS provide the greatest flexibility in both design and operations, while being afforded with some of the highest levels of fault tolerance. However, these very desirable features come at the cost of an overall increase in the system's complexity. Specifically, for a decentralized MRS to effectively operate in unstructured dynamic environments, the robotic system requires: (i) a distributed communication channel to share state variables and sensed data through an interaction network of a particular topology~\cite{A36}, possibly time-varying~\cite{A56}, (ii) a collective decentralized computing framework processing in real-time, data shared among units~\cite{C17}, and (iii) motion planning or collaborative control strategies, which are key to the effective coordination and division of labor among units having possibly different capabilities~\cite{burgard2005coordinated}. In practice, the divisions between these three key elements are not as clear-cut as it seems. On the contrary, MRS network architecture, distributed computing, and collective motion planning are profoundly intertwined, which is one of the reasons behind our incomplete understanding of biological swarming~\cite{B2}.

Once a communication channel is autonomously established without any external supporting infrastructure, the MRS requires situational awareness as it starts charting its unknown surrounding environment. This classically takes the form of a map, which has to be constructed and updated while a coordinated exploration strategy drives the system based on the current map information and new sensor data harvested by the various mobile units. Among the many different types of maps, the most common and intuitive are occupancy-grid maps (OGM), which are particularly befitting to MRS operations~\cite{birk2006merging}. Other types of maps include feature-based maps (e.g. line maps) and topological maps, which are significantly more computationally and memory efficient than OGM, yet require advanced sensor-data processing to generate an accurate and reliable map~\cite{rone2013mapping}.

As its name implies, OGM reduces a two-dimensional (2D) surface area into a grid of cells, which are characterized by several possible states---obstacle, unexplored, explored---identified through a probabilistic treatment of the noisy and uncertain sensor data. Range-sensors are particularly well-suited to the generation of OGM. For our purposes, given that we consider a swarm of miniature robots mapping dynamic environments, OGM constitute the optimal solution for three fundamental reasons. First, the smallness of our individual units requires effective obstacle sensing that can be miniaturized accordingly. Ultrasonic sensors have achieved small footprints, but their range is too short and are hardly as accurate as optical sensors. As is detailed below, our custom-made miniature LiDAR apparatus provides an effective-range sensing capability~\cite{C16}. Second, OGM being based on probabilistic estimations, they are intrinsically more tolerant to the presence of temporal changes in localized features, and thus lend themselves naturally to the mapping of dynamic environments. Third, the map building/updating process is almost identical when considering a single robot or a MRS, thereby conserving decentralization of the system's operations without compromising its scalability. The key challenge with OGM is its relatively high computational requirements when maintaining large grids~\cite{rone2013mapping}. However, this identified limitation can largely be alleviated when considering a decentralized computing framework as well as a distributed storage of the map. With a truly decentralized MRS---swarm robotics ones being one such particular type, individual units need only store and process a small subset of the sensed data and OGM. Effectively, not a single unit possesses the overall map, which should not be required for path-planning purposes.

The network architecture in MRS can either be centralized or decentralized. In the former, computation and control are performed by a single central entity, while in the latter, they are performed locally by the robots with minimal communication among the modules. The decentralized architecture is thus less prone to being affected by a single point of failure~\cite{rone2013mapping}. For this reason, and owing to recent technological developments, decentralized MRS architectures are getting prominence. For example in~\cite{djugash2008decentralized}, a decentralized algorithm was used to localize a flock of robotic sensor networks. Environmental monitoring tasks were performed by a decentralized swarm of robots in \cite{turduev2010cooperative,duarte2016application,A53}, including with heterogeneous swarms~\cite{C14}. Decentralized exploration and mapping of unknown indoor entity was demonstrated with a pair of autonomous quadcopters in~\cite{cesare2015multi}.

Despite the number of works on decentralized MRS with certain capabilities, it was found that existing MRS/swarm systems lack supporting experiments and performance data \cite{abukhalil2013comprehensive}, for example on how well they behave under varying number of robots (scalability), against the failure of the individuals (robustness), as well as in response to unknown and dynamic environments (flexibility), which are the key properties of swarm robotics~\cite{B2}. It was also observed in \cite{rone2013mapping} and \cite{saeedi2016multiple} that existing MRS were mostly applied to static environments and very few works have been done in the presence of dynamic circumstances. In terms of exploring unknown domains, it was noted in \cite{ozkil2011mapping} that while mapping of individual floors of an entity has been considered, simultaneous mapping of multi-floor territory has not been much addressed. This capability can add value to the overall mapping task. 

The ISR of modern urban environments, whose dense population navigate vertically as much as horizontally, requires of multistory exploration and monitoring~\cite{karg2010consistent}. There is no doubt that both tasks are extremely challenging, and particular so for fully autonomous and decentralized MRS. Beyond the dynamic nature of urban environments, there are three broad types of challenges associated with their autonomous exploration and surveillance: (1) intermittent accessibility to some spaces due to closed doors, (2) floor-to-floor transitions through staircases and/or elevators, and (3) ability to establish a communication channel in order to achieve inter-agent information transfer across wireless-hindering physical obstacles such as walls and floors. Attempts have been made to address these challenges separately: e.g. autonomous multi-floor exploration by a single robot~\cite{delmerico2013toward,delmerico2012ascending}, semi-autonomous exploration of multi-floor buildings with a legged robot~\cite{wenger2015semi}, navigation for service robots in the elevator environment~\cite{kang2007navigation}, autonomous multi-floor navigation by micro-aerial vehicles~\cite{shen2011autonomous}. However, to the best of our knowledge, the autonomous exploration and monitoring of an unknown indoor space consisting of multiple floors with dynamic features by a fully decentralized MRS remains an open challenge.

In this paper, a decentralized and heterogeneous system of custom-built miniature robots for autonomous exploration and mapping is developed in-house and evaluated to address those challenges facing current MRS. The performance of this swarming system in terms of its scalability, robustness, and flexibility is extensively tested  and quantitatively evaluated. The system is also then tested to simultaneously map two different storeys of considerable size within a campus environment, during normal operating hours and in the presence of students and staff. For the multi-floor mapping experiments, the large gap between the rooms/floors and the concrete walls/ceiling poses an operational challenge in terms of maintaining distributed communications required for our MRS to operate with robots on different floors. Intermediate communication relays located between the floors are necessary to ensure that a communication link is maintained between the swarms evolving on each floors. Indeed, having humans entering the area to physically place static nodes on the walls---especially in the case of hazardous environments---is neither desirable nor practical. This issue is overcome by having a heterogeneous MRS consisting of swarms of land robots teaming with multiple wall-climbing units. The wall-climbing units and ground-mapping ones share the same core architecture in terms of hardware and software; both variants have different sensory suites and mobility apparatuses. 


\section{System Architecture}

\subsection{Individual robotic unit: O-climb and O-map}
\begin{figure}[htb]
\centering
\includegraphics[width=\columnwidth]{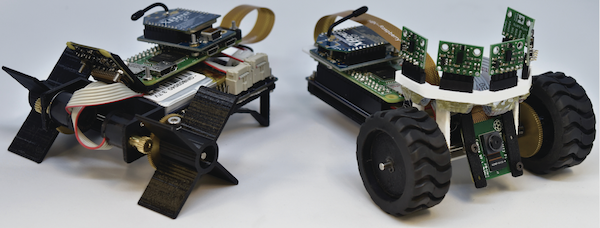}
\caption{O-climb (left) and O-map (right).}
\label{Orions}
\end{figure}

\begin{figure}[htb]
\centering
\includegraphics[width=\columnwidth]{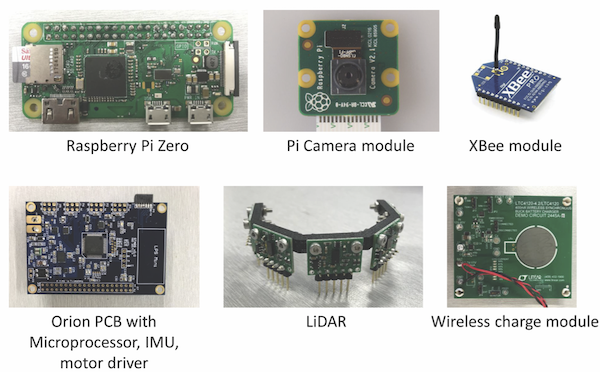}
\caption{The electronic suite of hardware components of the ORION platform. Modules on the bottom row---PCB, LiDAR, and wireless charger module---were designed and custom-built for ORION.}
\label{hardware}
\end{figure}

\begin{figure}[htb]
\centering
\includegraphics[width=\columnwidth]{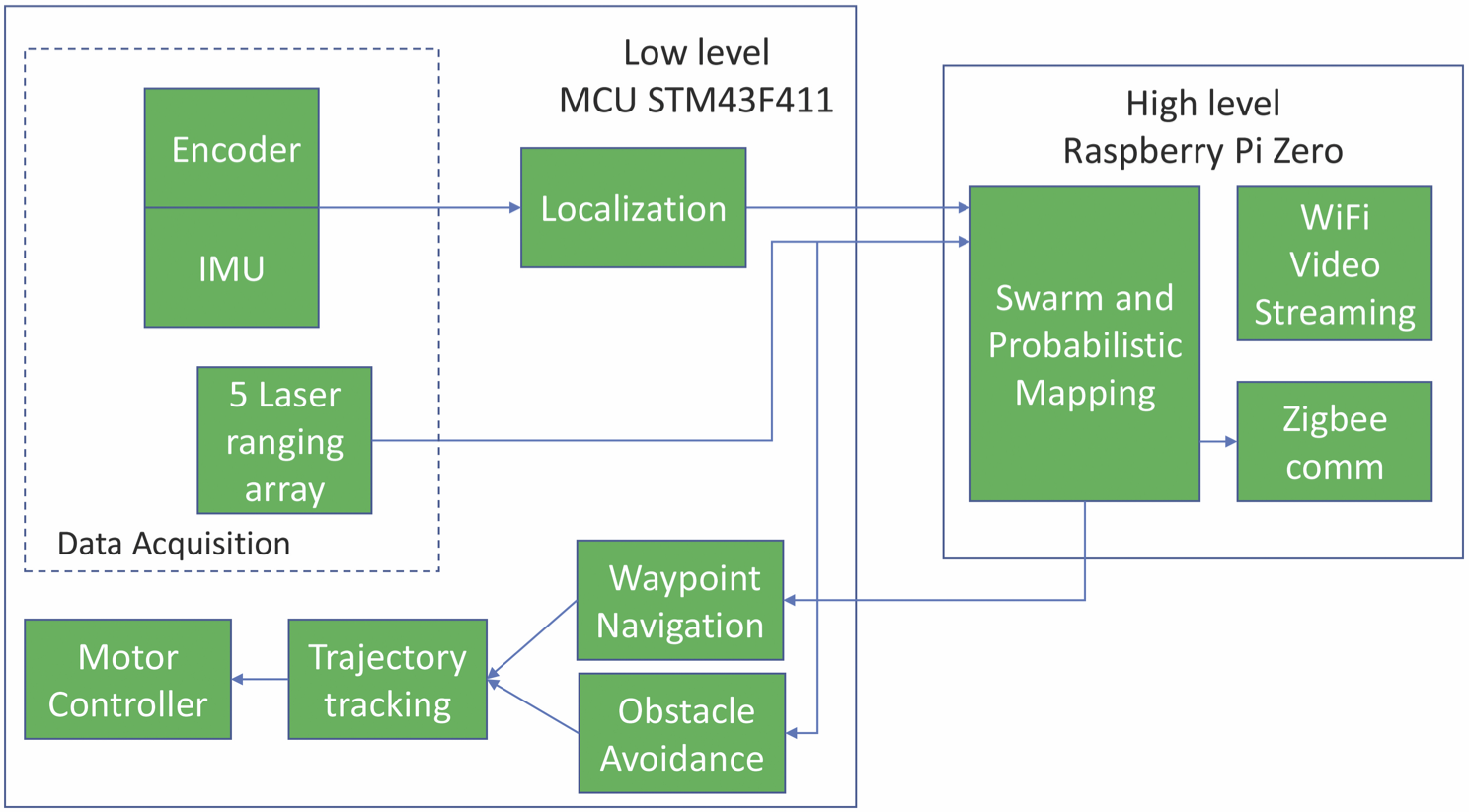}
\caption{Multi-block diagram of the developed software framework for ORION.}
\label{layers}
\end{figure}

Our MRS system is called ORION, and multiple generations and variants of ORION base units have been developed~\cite{C16,C15,C18,C20,A55}. Although this paper focuses on ORION's system-level design, features, and performance, we nonetheless provide in this section critical unit-level information and details about the two key variants: the ground-mapping (O-map) and the wall-climbing (O-climb) units, shown in Fig.~\ref{Orions}. Specifically, we report on the novel design of ORION's units as the main modules for the MRS and its unified system architecture. The hardware architecture of ORION can be mainly categorized into two parts: the chassis and the reconfigurable wheels. The chassis houses two DC motors and the electronics required for ISR as shown in Fig.~\ref{hardware} (see \cite{C15} for more details on the individual components). Sensors can be added or removed easily from the electronic architecture depending on the specific needs of a range of ISR tasks. The electronics stack is divided into two independent systems controlled by two software layers: the high-level and the low-level layers as shown in Fig.~\ref{layers}.

The high-level layer implements decentralized autonomous behaviors and performs the computationally-intensive tasks associated with them, such as computer vision, map-building, and path-planning. The low-level layer manages the platform sensors, actuators and power for real-time dynamic control, obstacle avoidance, and waypoint navigation. The two system layers communicate through a UART serial connection. The low-level layer constantly sends the sensors' data and robot's state to the high-level one for distributed autonomy processing. The high level, in turn, returns instructions at the low level for robot task execution.

With a common electronics stack on the chassis and reconfigurable wheels, the two design variants developed for the ORION MRS, each has its own set of unique capabilities: (i) O-map: a ground robot equipped with mapping function, live-video feed and distributed autonomy capabilities~\cite{C16}, and (ii) O-climb: a climbing robot equipped with live video feed and inter-floor communication capabilities~\cite{C15,C18,A55}, capable of robust internal and external transitions~\cite{C20}. As shown in Fig.~\ref{Orions}, O-map units (20 such individual units are operational) are equipped with rubber wheels and ball caster for ease of mobility on various terrains, while O-climb units (4 such individual units are in service) have special wheel-legs with compliant adhesive tape and a tail for robust climbing~\cite{C18,A55,C20,IFToMM}. The custom-built Light Detection and Ranging (LiDAR) sensor mounted on O-map~\cite{C16} is omitted in the O-climb variant as it does not perform mapping and to reduce its weight for ease of climbing. This shows the high level of modularity of the ORION's platform design for heterogeneous MRS purposes.

\subsection{Decentralized Swarming Technology}

While decentralization denies a MRS the advantages of a large core computational and communication hub, it offers three highly sought-after features. First, scalability: the system is capable of cooperatively operating under a wide range of system sizes.
Second, robustness: the MRS architecture allows for fault-tolerant operation when faced with sudden changes in its interaction network topology and the potential loss of multiple units.
Third, flexibility: with appropriate cooperative control strategies, the MRS is in principle capable of operating under unknown and dynamic circumstances. For effective decentralized MRS operations, the robots should be able to communicate in a distributed fashion establishing a dynamic (i.e. switching) communication network where nodes can be added or removed without compromising the MRS operations. To grant ORION these features, the robots are equipped with a swarm-enabling unit (SEU) based on the specifications detailed in~\cite{A40}. This SEU is platform-agnostic and has successfully been used in a swarm of land robots with varying interaction network topologies~\cite{A56} and in a heterogeneous swarm of surface vehicles performing a range of cooperative control strategies~\cite{C14,A53}.

Specifically, this is achieved with off-the-shelf, low-power and low-bandwidth wireless XBee-PRO modules capable of creating a distributed mesh network. The communication channel reconfigures itself automatically as the robots move and enter or leave each other's communication range. The performance of this distributed communication approach has been assessed with up to 40 robotic units with distances up to 160 meters and is reported in~\cite{A53}. This communication network is used by the robots to continuously broadcast their states---their estimated positions, sensor readings and current navigation waypoint---at a rate of 5 Hz. The maximum expected communication range is about 310 m in line of sight and the modules are capable of relaying messages through multiple hops in the network.

The decentralized swarming design principles and the cooperative control strategies are such that the successful collective operation of this system does not require a reliable global communication network between all the robots. The motion of each platform at a given time is determined solely by its own state and the current state of the neighboring robots. Therefore, only short-range and intermittent local communications are required.

\section{Collective Mapping Approach}

The decentralized ORION system of ground robots is designed to perform collective mapping operation in the absence of a central command.
This bars the possibility of system-level algorithms of coordination, global path-planning, or workload distribution among the individual units.
Instead, the collective mapping operation arises as emergent behavior from unit-level algorithms fed with information from neighboring agents.

Since our focus is on cooperative-control strategies for collective mapping, we do not consider the full problem of Cooperative Simultaneous Localization and Mapping (CoSLAM). We assume that all individual units know their initial poses in a common, global reference frame, and that a localization based on the unit's Inertia Measurement Unit (IMU) and the wheel encoder provides a sufficiently accurate positioning for the duration of the mapping task~\cite{C16}.
If higher accuracy is required, probabilistic localization algorithms such as unscented K\'alm\'an filter can be implemented as reported in~\cite{A40}.

\subsection{Map Representation}

ORION senses its surroundings by means of LiDAR readings that provide a point cloud of nearby obstacles.
This point cloud is then incorporated to a 2D OGM representation of the environment (see~\cite{C16} for more details).

The 2D environment is divided into a regular grid of cells with an associated state representing the posterior probability of occupancy.
Each square cell has a size of $1/15$~m and for simplicity all of them start with a prior occupancy probability of $0.5$.
With each LiDAR sensing, the probability of all the cells in the line-of-sight associated with the sensing are updated according to a Bayesian approach~\cite{thrun2003learning}. This classical OGM approach naturally lends itself to the mapping of dynamic environments since the state of a given cell can vary over time following the displacement of obstacles or the opening of a door.

Similar to the approach reported in~\cite{C17}, we consider Markovian processes whereby the robots only share their current LiDARs data---i.e. locally sensed information, not their entire local map. Formally, the local (individual) map $m$ of robot $i$ at time $t$ is obtained by computing the posterior probability $p(m|S_t^i)$ for a collection of sensory data $S_t^i$---its own sensed data and the sensed data from its connected neighbors identified through the interaction network adjacency $a_{ij}$ matrix---such that
\begin{equation}\label{eq_neighbors}
    S_t^i = \bigcup_{t'=1}^t\{s_j(t') ; j| a_{ij}(t')=1\},
\end{equation}
with $s_j(t')$ the sensed data of agent $j$ at instant $t'$, $a_{ii}(t)=1,\quad \forall t$, and $a_{ij}(t')=1$ if unit $i$ is connected with agent $j$ at instant $t'$, and $a_{ij}=0$ otherwise. It is worth stressing that this approach was designed to be compatible with temporal interaction networks, which have recently been found to be necessary to achieve effective collective responses by MRS~\cite{A56}.

\subsection{Communication Network}

The collective mapping operation requires the robots to exchange some key information with nearby neighbors such as the robot's pose, and LiDAR sensings.
As mentioned earlier, the onboard XBee modules form a wireless ad-hoc network, whose dynamic topology can be tuned to achieve optimal collective performance~\cite{A56}. Given the physical indoor environment that ORION operates in---i.e. no line-of-sight, presence of walls, low battery, latency and low-bandwidth constraints---it is expected that not all units are always connected.

In a system with units $A$, $B$ and $C$, if information from $A$ could not be sent to $C$, this does not suggests that $C$ will not have access to that information. Now, if $B$ gets access to the information from $A$ and includes it for its decision-making process, then the decision made by $B$, when observed by $C$, will have information from $A$ ``embedded" in it. This indirect propagation of behavioral information through the network is crucial for the collective mapping operation as it dictates how information will flow within the system. The work in~\cite{A56} reported the optimal network topologies for collective behaviors subjected to local perturbations, which, in the present framework, can be associated with moving obstacles/features.

\subsection{Exploration Strategy}

The swarm dynamics is based on a collective and distributed Frontier-Based Exploration strategy~\cite{yamauchi1998frontier}.
To decide where to move next, a robot computes at each instant a field of "preference potential" $V(\vec{r})$ representing a non-normalized probability of choosing a point $\vec{r}$ as the next waypoint.
The field $V$ is meant to prioritize the exploration of points that are: (i) near the frontier of the explored area, (ii) near the robot, and (iii) far away from the other units.
To account for this three factors, we have defined the field as
\begin{equation}
    V_i(\vec{r}) = V_F(\vec{r}) \times
     \frac{1}{\min(\|\vec{r} - \vec{r_i}\|, R_0)} \times
     \prod_{j\sim i} \|\vec{r} - \vec{r_j}\|^2 \, ,
\end{equation}
where $R_0>0$ is an arbitrary cut-off distance, $V_F$ is a term characterizing the frontier between the explored free space and the unexplored space, and $j\sim i$ represents the set of neighbors $j$ of unit $i$. This frontier is obtained by applying a classical Roberts cross operator to the Bayesian map. The waypoint is then selected as the point that maximizes $V(\vec{r})$. This collective mapping approach is formalized in Algorithm~\ref{alg:cm}.
\begin{algorithm}
\caption{Collective Mapping algorithm}\label{alg:cm}
\begin{algorithmic}[1]
\Procedure{CollectiveMapping}{$r,R,s,map$}
\State $map\gets UpdateOccupancyGrid(map,s)$
\State $neighbors \gets ConnectedNeighbors(r,R)$
\For{$n$ in $neighbors$}
\State $map\gets UpdateOccupancyGrid(map,s_{n})$
\EndFor
\State $waypoint \gets FrontierExploration(map,r,R)$
\State \textbf{return} $map, waypoint$
\EndProcedure
\end{algorithmic}
\end{algorithm}

\section{Experimental Results \& Discussion}
Here, we present the results of multiple experiments characterizing and quantifying some key features of our decentralized MRS ORION.

\subsection{Scalability experiments}
The first series of tests assess how the performance of the collective mapping operations scales with the MRS size $N$.
An unknown, irregular floor space (see video~\cite{yt}) of approximately 60 m$^2$ with 5 irregular obstacles in it has been used for all scalability tests.
The corresponding surface area is more than $4000$ times the footprint of a single unit ($12 \times 12$~cm). Clearly, increasing the number of robots beyond $6$ yields only marginal improvement to the system's performance. Given the size of the surface area used, operating beyond $N=8$ would inevitably lead to drops in performance owing to the overcrowding of the arena. However, this does not mean that our system is not capable of operating with $N>8$ units, quite the contrary (see Sec.~\ref{Multi-floor exploration}).

\begin{figure}[htb]
\centering
\includegraphics[width=\columnwidth]{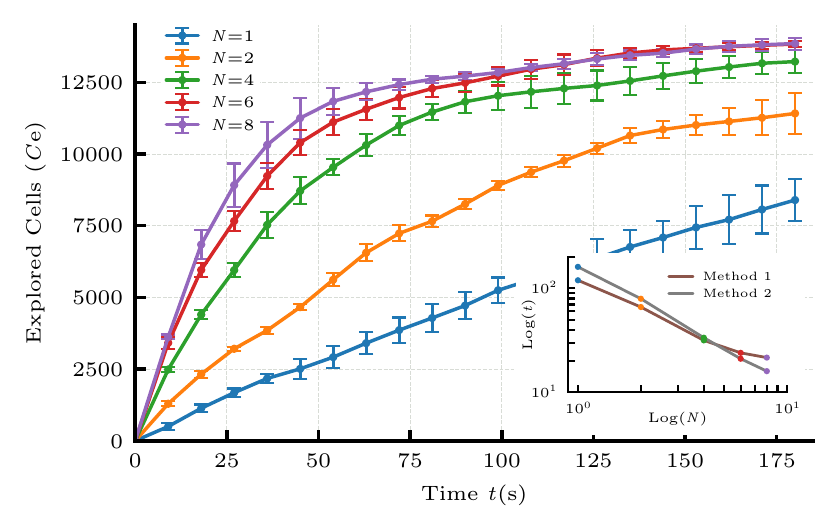}
\caption{Scalability experiments for five different MRS sizes: $N=1,2,4, 6$ and $8$. Each point---mean value and the associated standard deviation---is obtained from the results of five repeated experiments with same $N$ value and identical initial conditions. The insert shows the scaling of two characteristic times with $N$ for these collective mapping operations by means of Methods 1 and 2.}
\label{scalability}
\end{figure}

The results, presented in Fig.~\ref{scalability}, reveal a rapid initial increase in the number of  explored cells $C_e$ that subsequently reaches a plateau for large $N$. The initial fast increase in $C_e$ is to be expected the whole space is unknown, hence all cells encountered by the units are unexplored. At a later stage, the coordination among robots imposes them to travel farther in search of new unexplored cells. We post-processed these data to estimate the characteristic time-scale of collective mapping in the early stages (Method 1) and at a later stage when $C_e=8,000$. Specifically, Method 1 assumes that $C_e$ progresses according to $C_e(t)=C_e^{\infty}(1-\exp(-t/\tau))$ and estimates the time-scale of operation $\tau$ by means of linear regressions of $\log (C_e/C_e^\infty)$ for each value of $N$. Method 2 gives the time it takes to explore the first $8,000$ cells. Interestingly, both time-scales exhibit an almost identical linear scaling (in a log-log plot) with the system size $N$ as shown in the insert of Fig.~\ref{scalability}.

\subsection{Robustness and recovery experiments}

As a next step, the fault-tolerance of ORION is assessed by forcing the removal of 2 units out of a total of $N=4$ in the early stages of the collective mapping process of the same floor area as in the scalability experiments. Specifically, at $t=20$~s, two units are manually switched off thereby leaving the two remaining units operating by themselves. At $t=120$~s, two ``new" units are injected into the pen and instantly start swarming with the other two units. Beyond the characterization of robustness, this series of experiments allow us to analyze the dynamics of the emergent recovery process.

\begin{figure}[htbp]
\centering
\includegraphics[width=\columnwidth]{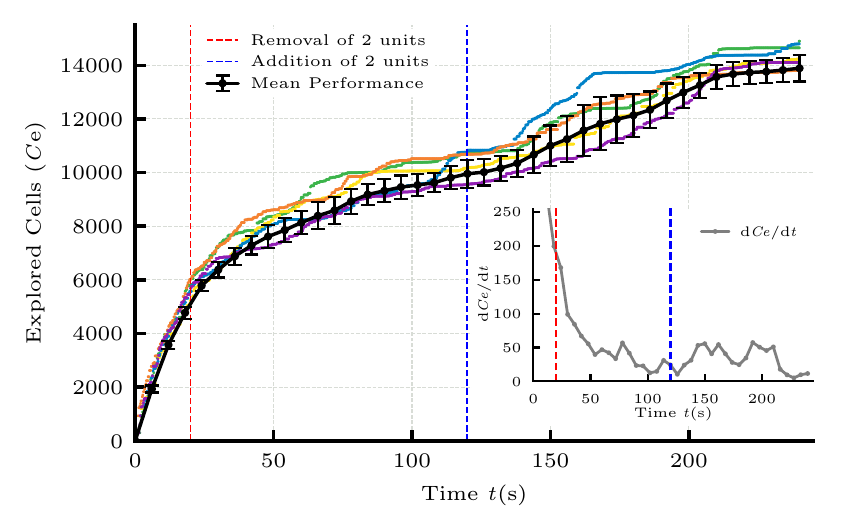}
\caption{Robustness experiments: each of the five experiments are shown with different colored dots. The mean performance and its associated standard deviation is shown in black. The red (resp. blue) dashed lines correspond to the instant at which two units are removed (resp. added) of the MRS. The insert shows the rate of change of the number of explored cells $C_e$.}
\label{robustness}
\end{figure}

Figure~\ref{robustness} presents the results of five successive experiments (colored dots) as well as the mean performance (with standard deviation in black color). When the removal of the two units takes place, there is a noticeable drop in performance as the number of explored cells slows down. Hundred seconds later, when two new units are injected, the rate at which $C_e$ increases also increases but only after a 20-30 seconds delay (the rate of change of $C_e$ with time is shown in the insert of Fig.~\ref{robustness}). This delay can be justified by the time needed by the newly formed system to achieve an effective coordinated behavior. The two present units may see their direction of travel affected by the LiDAR information from the two added units. However, passed this delay, the 4-unit MRS exhibits a clear improvement in its performance as attested by the increase in $dC_e/dt$ shown in the insert of Fig.~\ref{robustness}.

In real-world scenarios, the removal of units occurs with hardware breakdowns, low battery levels, and most commonly with out-of-range communications. Our previous work~\cite{A53} provides a statistical study on communication success based on distance apart for varying sizes of the MRS.

\subsection{Flexibility experiments}

Here, we consider the collective response of the system to a drastic increase in the area to explore. Specifically, a rectangular ($3\times 9$~m) $27$~m$^2$ space is divided into three square areas of almost identical surface area of $9$~m$^2$ (see video~\cite{yt}). Initially, two robots are placed in the rightmost area, and another two robots are started from the leftmost area. The central area is empty and inaccessible owing to the presence of two walls separating it from the leftmost and rightmost areas. At $t=0$, each pair of robots starts mapping its exclusive area. But after $30$~s, the two walls are removed, thereby opening the two areas mapped by the pairs of O-map to a central void.

\begin{figure}[htbp]
\centering
\includegraphics[width=\columnwidth]{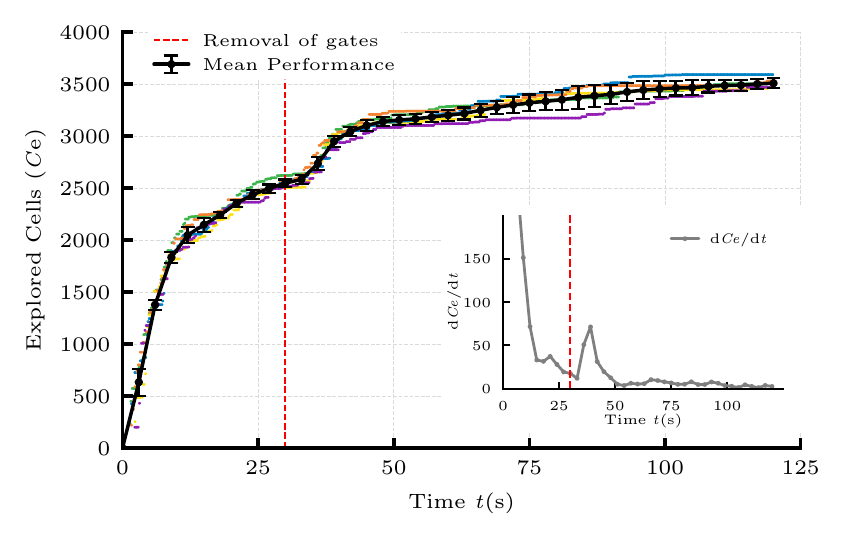}
\caption{Flexibility experiments: each of the five experiments are shown with different colored dots. The mean performance and its associated standard deviation is shown in black. The red dashed lines correspond to the instant at which the floor area is expanded by 50\%. The insert shows the rate of change of the number of explored cells $C_e$.}
\label{flexibility}
\end{figure}

The results for five distinct experiments are reported in Fig.~\ref{flexibility}, showing a high consistency in performance. Before the removal of the two walls, the performance is analogous to the one obtained for our scalability experiments. However, once the walls are removed and the area to explore expanded, a change in performance can clearly be detected after a few seconds delay. Again, this observed delay can be associated with the time it takes to the MRS to adjust its coordinated behavior to changing circumstances. The sharp increase in $dC_e/dt$ shown in the insert of Fig.~\ref{flexibility} after ther removal of the walls is a good indicator. It is therefore very similar to the collective behavior observed during the recovery experiment discussed previously.

\subsection{Multi-floor exploration} \label{Multi-floor exploration}

\begin{figure}[htb]
\centering
\includegraphics[width=\columnwidth]{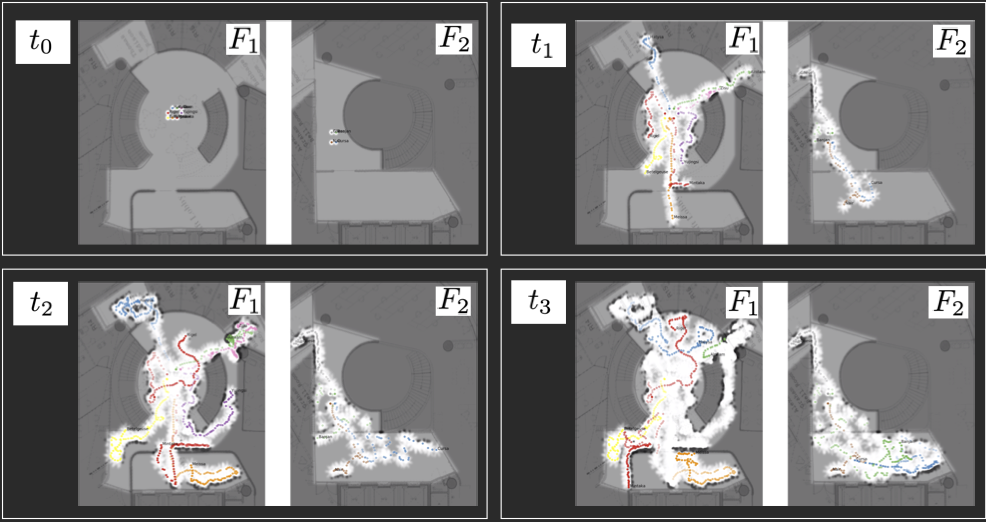}
\caption{Snapshots at successive time instants $(t_0,t_1,t_2,t_3)$ of the mapped areas by 12 units across two floors ($F_1$ and $F_2$).}
\label{two-floors}
\end{figure}

As a last step, we present some qualitative results of a collective exploration over two floors. These results were obtained on our university campus grounds during operating hours. The areas mapped constitute unstructured and dynamic environments, and no external/additional supporting infrastructure was necessary to run the experiments. Eight O-map units were initially positioned on one floor ($F_1$), and four others on the floor above ($F_2$). Two O-climb units were placed on a vertical wall and ascended during the experiments, thereby expanding the distributed communication network between floors. The layouts of the surface areas to be mapped on $F_1$ and $F_2$ are shown in light gray in Fig.~\ref{two-floors}: they consist of open spaces, elevator lobbies and office rooms with open doors. The overall collective exploration task lasted about 3 to 4 minutes depending on the dynamic obstacles encountered. Snippets of the live operations are shown in the final segment of the video~\cite{yt}.

\section{Conclusions}
In this paper, we presented the analysis, design, and development of a decentralized multi-robot system, ORION, capable of performing scalable and fault-tolerant explorations of unknown multi-floor indoor environments. ORION operates based on a swarm robotic subsystem that is collectively mapping possibly dynamic environments made up of multiple floors. The wall-climbing units achieve wall-adhesion by means of dry-adhesive whegs and share the same basic robotic architecture as the mapping variant. They serve to expand the distributed communication network between the swarming ground robots that are mapping two floors.

ORION exhibits consistent scalability, robustness and flexibility when collectively mapping unknown and unstructured floor spaces with several irregular obstacles present. It is also capable of recovery by subsequently injecting two other robots while it is collectively operating. Our results also reveal a delayed collective response in two particular cases: (1) when the MRS is rapidly expanded by the addition of several units, and (2) when the MRS responds to abrupt changes in the topology of the environment being mapped.



\end{document}